\documentclass[sigconf]{acmart}

\usepackage{CJKutf8}
\usepackage{algorithm}
\usepackage{subfigure}
\usepackage{xcolor}
\usepackage{booktabs} 
\usepackage{mathtools}
\usepackage{amsmath,bm}
\usepackage{makecell}
\usepackage{multirow}
\usepackage{multicol}
\usepackage{array}
\usepackage{comment}
\usepackage{textcomp}
\usepackage{fmtcount}
\usepackage{flushend}

\newcommand{\tabincell}[2]{\begin{tabular}{@{}#1@{}}#2\end{tabular}}

\usepackage[noend]{algpseudocode}
\usepackage{algorithmicx,algorithm}
\AtBeginDocument{%
  \providecommand\BibTeX{{%
    \normalfont B\kern-0.5em{\scshape i\kern-0.25em b}\kern-0.8em\TeX}}}

\setcopyright{acmcopyright}

\copyrightyear{2021} 
\acmYear{2021} 
\setcopyright{acmcopyright}\acmConference[KDD '21]{Proceedings of the 27th ACM SIGKDD Conference on Knowledge Discovery and Data Mining}{August 14--18, 2021}{Virtual Event, Singapore}
\acmBooktitle{Proceedings of the 27th ACM SIGKDD Conference on Knowledge Discovery and Data Mining (KDD '21), August 14--18, 2021, Virtual Event, Singapore}
\acmPrice{15.00}
\acmDOI{10.1145/3447548.3467057}
\acmISBN{978-1-4503-8332-5/21/08}

\begin{document}

\fancyhead{}

\title{AliCG: Fine-grained and Evolvable Conceptual Graph
Construction for Semantic Search at Alibaba}

 \renewcommand{\shorttitle}{AliCG: Fine-grained and Evolvable Conceptual Graph
Construction}


\author{
Ningyu Zhang$^{1,2*}$, 
Qianghuai Jia$^{4*}$, 
Shumin Deng$^{1,2*}$, 
Xiang Chen$^{1,2}$, 
Hongbin Ye$^{1,2}$, 
Hui Chen$^{3}$, 
Huaixiao Tou$^{3}$, 
Gang Huang$^{5}$, 
Zhao Wang$^{1}$, 
Nengwei Hua$^{3}$, 
Huajun Chen$^{1,2\star}$
}

\affiliation{
$^1$Zhejiang University\country{China} \& AZFT Joint Lab for Knowledge Engine\country{China}
$^3$Alibaba Group\country{China}
}
\affiliation{
$^2$Hangzhou Innovation Center\country{China}, Zhejiang University\country{China},
$^4$AntGroup\country{China}
$^5$Zhejiang Lab\country{China}
}

\email{
{zhangningyu, 231sm, xiang_chen, yehongbin, huanggang, zhao_wang, huajunsir}@zju.edu.cn, 
}
\email{
{qianghuai.jqh, weidu.ch, huaixiao.thx, nengwei.huanw}@alibaba-inc.com
}



 \renewcommand{\shortauthors}{Ningyu Zhang et al.}

\begin{abstract}
Conceptual graphs, which is a particular type of Knowledge Graphs, play an essential role in semantic search. Prior conceptual graph construction approaches typically extract high-frequent, coarse-grained, and time-invariant concepts from formal texts. In real applications, however, it is necessary to extract less-frequent, fine-grained, and time-varying conceptual knowledge and build taxonomy in an evolving manner.  In this paper, we introduce an approach to implementing and deploying the conceptual graph at Alibaba. Specifically, We propose a framework called \textbf{AliCG} which is capable of a) extracting fine-grained concepts by a novel bootstrapping with alignment consensus approach, b) mining long-tail concepts with a novel low-resource phrase mining approach, c) updating the graph dynamically via a concept distribution estimation method based on implicit and explicit user behaviors. We have deployed the framework at Alibaba UC Browser. Extensive offline evaluation as well as online A/B testing demonstrate the efficacy of our approach.

\noindent\let\thefootnote\relax\footnotetext{
$*$ Equal contribution and shared co-first authorship. \\
$\star$ Corresponding author.
}
\end{abstract}

\ccsdesc[500]{Information systems~Query representation}
\ccsdesc[500]{Information systems~Information extraction}

\keywords{Concept Mining; Taxonomy Construction; Knowledge Graph}

\maketitle

\section{Introduction}
Knowledge is important for text-related applications such as semantic search.  Knowledge Graphs (KGs) organize facts in a structured graph way as triples in the form of \emph{<subject, predicate, object>}, abridged as $(s, p, o)$, where $s$ and $o$ denote entities and $p$ builds relations between entities.  \textbf{Conceptual Graph}, which is  a special type of KGs, builds the  semantic connections between concepts and has proven to be valuable in \emph{short text understanding} \cite{wang2014concept}, \emph{Word Sense Disambiguation} \cite{suzuki2018all}, \emph{enhanced entity linking} \cite{chen2018short}, \emph{semantic query rewriting} \cite{wang2015query}, etc. Essentially, conceptualization helps humans generalize previously gained knowledge and experience to new settings, which may reveal paths to high-level cognitive \textbf{System 2} \cite{bengio2017consciousness} in a conscious manner.   In real-life applications, the conceptual graph provides valuable knowledge to support many applications \cite{zhang2021drop}, such as semantic search.  Web search engines (e.g., Google and Bing) leverage a taxonomy to better understand user queries and improve the search quality.  Moreover,  many online retailers (e.g., Alibaba and Amazon) organize products into categories of different granularities so that customers can easily search and navigate this category taxonomy to find the items they want to purchase. 

 \begin{figure*}
\centering
\includegraphics [width=1\textwidth]{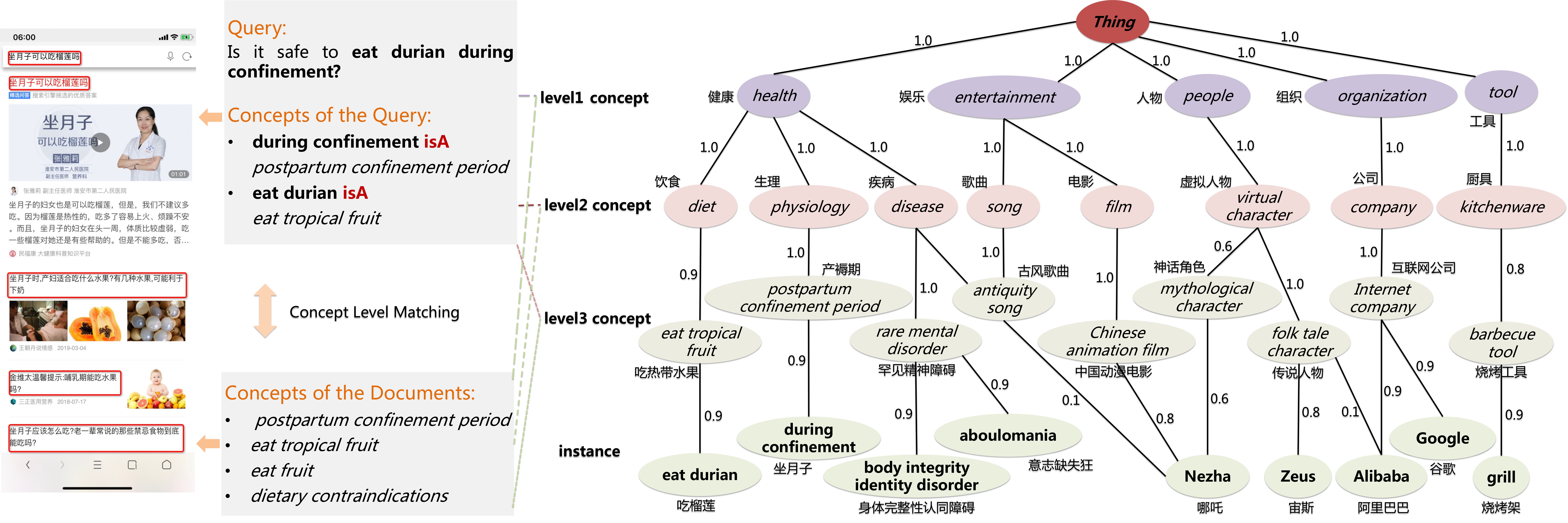}
\caption{Data hierarchy of Alibaba Conceptual Graph (AliCG) for semantic search.}
\label{alicg}
\end{figure*}

In this paper, we introduce the Alibaba Conceptual Graph (AliCG), which is a large-scale conceptual graph of more than 5,000,000 fine-grained concepts, still in fast growth, automatically extracted from noisy search logs. As shown in Figure \ref{alicg}, AliCG comprises four levels: \textbf{level-1} consists of concepts expressing the domain that those instances belong to;  \textbf{level-2} consists of concepts referred to the type or subclass of instances;  \textbf{level-3} consists of concepts that are the fine-grained conceptualization of instances expressing the implicit user intentions;  \textbf{instance layer} includes all instances such as entities and non-entity phrases.  AliCG is currently deployed at Alibaba to support a variety of business scenarios, including the product Alibaba UC Browser.  AliCG has been applied to more than dozens of applications in Alibaba UC Browser, including intent classification, named entity recognition, query rewriting, and so on, and it receives more than 300 million requests per day. 

Building AliCG is not a trivial task. Previous studies such as YAGO \cite{rebele2016yago} and DBPedia \cite{auer2007dbpedia} have investigated the extraction of knowledge from formal texts (e.g., Wikipedia). Probase \cite{wu2012probase} proposes an approach for extracting concepts from semi-structured Web documents. However, these approaches could not be adapted to our applications because several challenges remain unresolved.
 
\textbf{Fine-grained Concept Acquisition.}  Conventional approaches devoted to extracting coarse-grained concepts such as categories or types.  However, in Alibaba's scenario of question  fine-grained concepts are necessary to increase the recall of answer results.  For example,  “grill (\begin{CJK}{UTF8}{gbsn}烤架\end{CJK})” is a “tool (\begin{CJK}{UTF8}{gbsn}工具\end{CJK})” and “scarf (\begin{CJK}{UTF8}{gbsn}围巾\end{CJK})” is a "clothes (\begin{CJK}{UTF8}{gbsn}服饰\end{CJK})”. However, it would be more helpful if we can infer that a user searching for these items may be more interested in “barbecue tool (\begin{CJK}{UTF8}{gbsn}烧烤工具\end{CJK})” or "keep warm clothes  (\begin{CJK}{UTF8}{gbsn}保暖服饰\end{CJK})” rather than another "tool (\begin{CJK}{UTF8}{gbsn}工具\end{CJK})"  like “wrench (\begin{CJK}{UTF8}{gbsn}扳手\end{CJK})”—these concepts are rare in existing conceptual graphs.  

\textbf{Long-tail  Concept Mining.} Conventional approaches  \cite{wu2012probase} generally extract concepts based on Hearst patterns, e.g., "especially" and "such as." However, these approaches cannot extract long-tail concepts from extremely short or noisy queries, which are common in search engines. For instance, it is non-trivial to extract the concept "rare mental disorder (\begin{CJK}{UTF8}{gbsn}罕见精神障碍\end{CJK})" of the instance "body integrity identity disorder (\begin{CJK}{UTF8}{gbsn}身体完整性认同障碍\end{CJK})" from the search log as only 35 instances mentioned "rare mental disorder". It is rather difficult to extract such concepts from the short text with pattern matching (the pattern is too general \cite{zhang2019gcn,DBLP:conf/wsdm/DengZKZZC20,DBLP:conf/www/ZhangDSCZC20,DBLP:journals/corr/abs-2104-07650,ACL2021_OntoED}, and there is little context information as well as co-occurrence samples), as Figure \ref{arc1} shows. Besides, there exist lots of scattered concepts in user search engines such as "traditional activities Tibetan New Year (\begin{CJK}{UTF8}{gbsn}藏历新年习俗\end{CJK})". Recent approaches usually regard such concept extraction procedure as sequence labeling tasks \cite{liu2019user}, which rely on a tremendous amount of training data for each concept.  Nearly 79\% of the concepts are long-tail in the search logs. Therefore, it is crucial to be able to extract concepts with limited numbers of instances.   

\textbf{Taxonomy Evolution.} Numerous instances and concepts in user search queries are related to recent trending and evolving events. Conventional approaches are not able to update the taxonomies over time. For instance, a user may search "Nezha (\begin{CJK}{UTF8}{gbsn}哪吒\end{CJK})" or "New animations in February  (\begin{CJK}{UTF8}{gbsn}二月新番)\end{CJK}" in  search engines.  The implied meaning of such concepts changes over time because apparently, "Nezha" has different meanings at different times (e.g., Chinese animation film, mythological character, the hero in Honor of Kings), and there are various new animations in February in different years.  Thus, it is imperative to incorporate the temporal evolution into the taxonomies.  However, as we extract instances and concepts from the text, which inevitably brings about duplicate edges, it is necessary to align those nodes with the same meaning in the conceptual graph. Besides, as there are many multiple-to-multiple nodes in the conceptual graph, it is prohibitive to update such complex graphs over time.  In other words, it is difficult to estimate the confidence distribution of the concepts given instances.   

To address challenges mentioned above, we propose the following  contributions in the design of AliCG: 
 
\begin{figure*}
\includegraphics [width=1\textwidth]{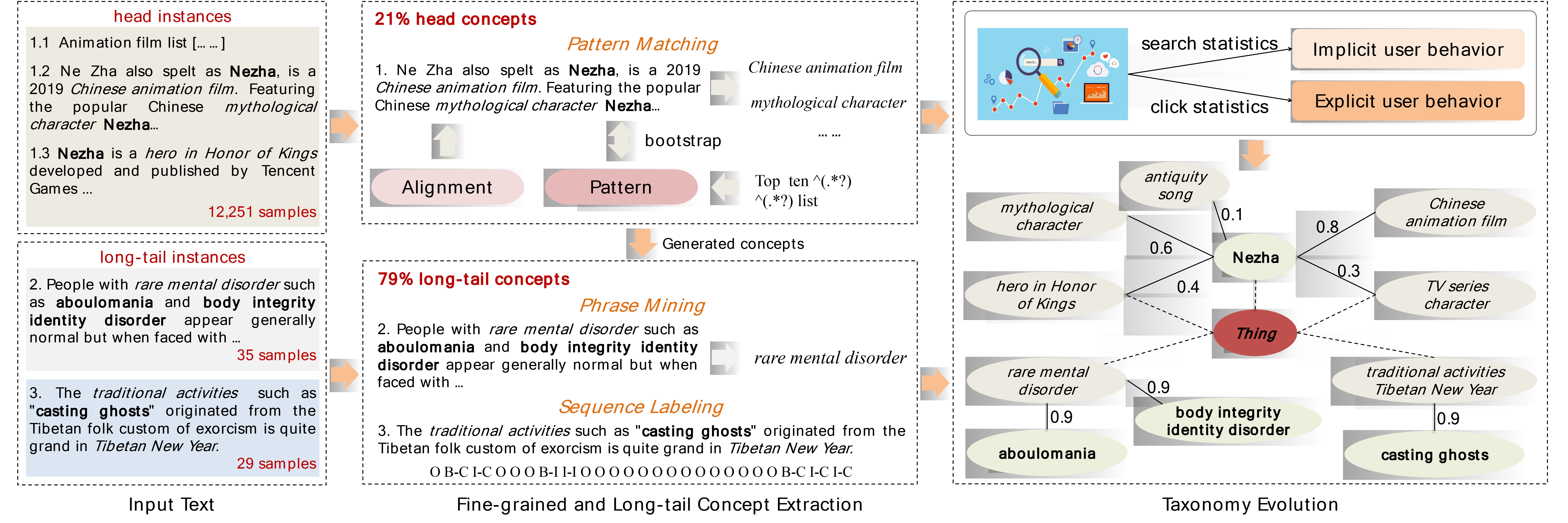}
\caption{Framework of Alibaba  Conceptual Graph  Construction.}
\label{arc1}
\end{figure*}

\emph{First}, we propose a novel  \textbf{\emph{bootstrapping with the alignment consensus}} approach to tackle the first challenge of extracting fine-grained concepts from noisy search logs. Specifically, we utilize a small number of predefined string patterns to extract concepts, which are then used to expand the pool of such patterns. Further, the new mined concepts are verified with query-title alignment; that is, an essential concept in a query should repeat several times in the document title frequently clicked by the user.   \emph{Second}, we introduce a novel \textbf{\emph{conceptualized phrase mining and self-training with an ensemble consensus}} approach to extract long-tail concepts. On the one hand, we extend the off-the-shelf phrase mining algorithm with conceptualized features to mine concepts unsupervisedly.  On the other hand, we propose a novel low-resource sequence tagging framework, namely, self-training with an ensemble consensus, to extract those scattered concepts.  \emph{Finally},  we propose a novel \textbf{\emph{concept distribution estimation method based on implicit and explicit user behaviors}} to tackle the taxonomy evolution challenges. We employ concept alignment and take advantage of user's searching and clicking behaviors to estimate the implicit and explicit concept distributions to construct a four-layered concept--instance taxonomy in an evolving manner. 

To deploy AliCG in real-life applications, we further introduce three methods for utilizing the conceptual knowledge, including, \textbf{\emph{text rewriting}}, \textbf{\emph{concept embedding}}, and \textbf{\emph{conceptualized pretraining}}.  We conduct extensive offline evaluations, including concept mining and applications such as intent classification and named entity recognition. Experimental results show that our approach can extract more fine-grained concepts in both normal and long-tail setting compared with baselines. Moreover, the performance of intent classification and named entity recognition is significantly improved when integrated with the conceptual graph.  We also perform online  A/B testing on more than 400 million actual users of the Alibaba UC Browser mobile application\footnote{\url{https://www.ucweb.com/}}.  The experimental results show that the relevant score increases by 12\%, the click-through rate (CTR) increase by 5.1\%, and page view (PV) increases by 5.5\%. The highlights of our work  are the following: 
 \begin{itemize}
\item We introduce the  AliCG, which is a large-scale conceptual graph of more than 5,000,000 fine-grained concepts automatically extracted from the noisy search logs\footnote{Demo available in \url{http://openconcepts.zjukg.cn/}}.
\item  We propose three novel approaches to address the issues of fine-grained concept acquisition, long-tail concept mining, and taxonomy evolution. Our framework is able to extract and dynamically update concept taxonomy in both normal and long-tail settings. 
\item  We introduce three methods to deploy our approach in Alibaba UC Browser, and both offline evaluation and online A/B testing demonstrate the efficacy of our approach.   We also release an open dataset\footnote{\url{https://github.com/alibaba-research/ConceptGraph}} of AliCG with 490,000 instances for research purposes.    
\end{itemize}

\section{Methodology}
Our approach is aimed at constructing conceptual graph from both the Web documents and the noisy query logs.  We denote a user query by $q=w_{1}^{q} w_{2}^{q} \cdots w_{|q|}^{q}$ and the set of all queries by $Q$. In addition, we denote a document  by $d=w_{1}^{d} w_{2}^{d} \cdots w_{|d|}^{d}$. Given a user query $q$, and  the top-ranked clicked documents, $D^q = \left\{d_{1}^{q}, d_{2}^{\hat{q}}, \cdots, d_{\left|D^{q}\right|}^{q}\right\}$, we aim to extract an instance/concept phrase, $\mathbf{c}=w_{1}^{c} w_{2}^{c} \cdots w_{|\mathbf{c}|}^{c}$. As illustrated in Figure \ref{arc1}, the construction of the conceptual graph generally consists of three modules, as follows:

\textbf{Fine-grained Concept Acquisition} aims at extracting those common fine-grained concepts from the noisy search logs with    \textbf{\emph{bootstrapping with the alignment consensus}}. After that, candidate concepts and instances are acquired, and we link those instances with corresponding concepts via probabilistic inference and concept matching following the approach proposed in \cite{liu2019user}. 

\textbf{Long-tail Concept Mining} aims at extracting those long-tail concepts from the noisy search logs with \textbf{\emph{conceptualized phrase mining and self-training with an ensemble consensus}}. In addition, we also leverage concepts from the fine-grained concept acquisition module as distantly supervised samples for concept mining (see details in section \ref{sec-long-tail}).

\textbf{Taxonomy Evolution}  aims at building the taxonomy in an evolving way with \textbf{\emph{concept distribution estimation method based on implicit and explicit user behaviors}}. Note that taxonomy evolution is conducted with those concepts mined from the modules of fined-grained concept acquisition and long-tail concept mining. 

\subsection{Fine-grained Concept Acquisition\label{link}}
Previous studies \cite{wu2012probase} show that the iterative (bootstrapping) approaches can extract coarse isA facts with the highest confidence starting with a set of seed patterns; thus, it is intuitive to construct elegant patterns to obtain fine-grained concepts.  Specifically, we define a small set of patterns to extract concept phrases from queries with high confidence. For example, “Top 10 XXX (\begin{CJK}{UTF8}{gbsn}十大XXX)\end{CJK}” is a pattern  that can be used to extract seed concepts. Based on this pattern, we can extract concepts such as "Top 10 mobile games (\begin{CJK}{UTF8}{gbsn}十大手游)\end{CJK}." However, there still exist lots of noisy texts, which are challenging for vanilla bootstrapping approaches.  To this end, a novel bootstrapping with the alignment consensus approach is proposed to deal with noisy texts. The intuition behind this is that \emph{we must control the pattern generalization and concept consistency given query-title pairs}. 

\textbf{Bootstrapping with Alignment Consensus.} First, an extracted pattern should not be too general for the sake of extracting \emph{fine-grained concepts}. Therefore, given a new pattern $p$ found in a certain round, let $n_s$ be the number of concepts in the existing seed concept set that can be extracted by $p$ from the query set $Q$. Let $n_e$ be the number of new concepts that can be extracted by $p$ from $Q$. We keep pattern $p$ via  the function $Filter(p)$: 1) $\alpha<\frac{n_{s}}{n_{e}}<\beta$, and 2) $n_{s}>\delta$, where $\alpha$, $\beta$, and $\delta$ are predefined thresholds to control the fineness of extracted concepts.   Second, we filter the mined concepts via query-title alignment to improve the quality of fine-grained concepts.  Even though bootstrapping helps discover new patterns and concepts from the query set $Q$ in an iterative manner, such a pattern-based method has limited extraction ability and may introduce considerable noise. Based on \cite{liu2019user}, we further propose the extraction of concepts from a query and its top clicked link titles in the search log as the essential concept in a query should repeat several times in the document title frequently clicked by the user. The overall algorithm is shown below:

\begin{algorithm}[th]
\begin{algorithmic}[1]
\caption{Bootstrapping with the alignment consensus} 
\Require query and high-clicked documents set $ (q,D) \in T$.
\Require Predefined templates set $p \in P$, Concepts $C^{head} = \Phi$
\For{\emph{iter} iterations}
\For{each $p \in R$}
\For{each $(q,D)\in T$}
\If{$q$ match $r$}
\State $c^p = ExtractByTemplate(q)$
\State $c^a = ExtractByAlignment(q,D)$
\If{$len(c^p) <= len(c^a)$}
\State $C^{head}  \leftarrow C^{head}  + c^p$
\EndIf
\EndIf
\EndFor
\EndFor
\State $p^{candidate} = GenPattern(C^{head},q)$
\State $p^{new} = Filter(p^{candidate})$
\State $P \leftarrow P + p^{new}$
\EndFor
\Return $C^{head}$ 
\label{bootstrap} 
\end{algorithmic}
\end{algorithm}

Note that we have mined typical fine-grained instance and concept candidates, but we should link those instances with corresponding concepts. Firstly, we utilize a concept discriminator to determine whether each candidate is a concept or instance.  We represent each candidate by a variety of features, such as whether this concept has ever appeared as a query, how many times it has been searched, etc. We then train a classifier with  Gradient Boosting Decision Tree and link instances\footnote{We also mined instances from the search logs based on keyword extraction to increase the coverage of instance.} with those classified concepts by probabilistic inference and concept matching following \cite{liu2019user}. 

\subsection{Long-tail Concept Mining\label{sec-long-tail}}
 
\begin{figure}
\centering
\includegraphics [width=0.4\textwidth]{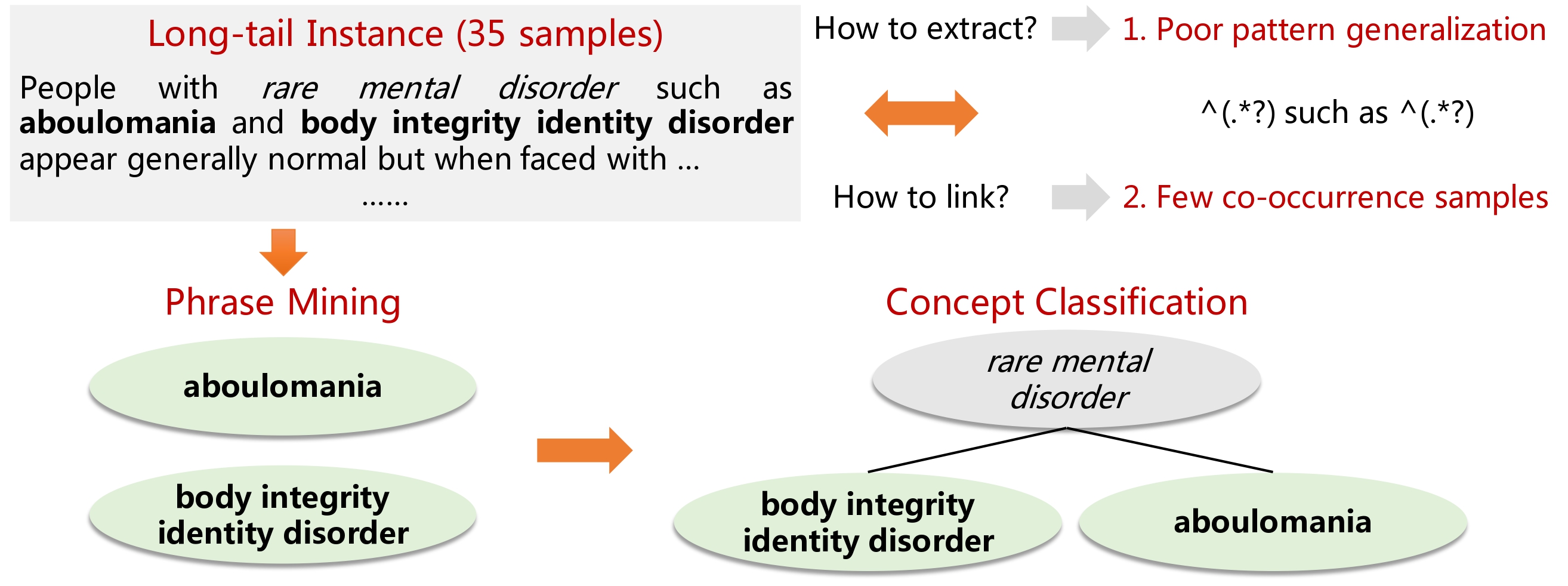}
\caption{Long-tail Concept Mining.}
\label{long_tail_concept_mining}
\end{figure}

Although iterative pattern matching can extract lots of high-frequent concepts, it is still non-trivial to extract those with only a few instances. As Figure \ref{long_tail_concept_mining}  shows, it is challenging to extract long-tail concepts because of two main reasons:  \emph{poor pattern generalization}  and  \emph{few co-occurrence samples}. Thus, it is difficult to match such low-shot concepts and link those concepts with corresponding instances. For the long-tail problem, we first propose \emph{conceptualized phrase mining} to utilize the external domain knowledge graph to generate weak-supervised samples for unsupervised learning of those long-tail concepts, which address those issues.  We then propose a self-training-based low-resource tagging algorithm for supervised learning of long-tail concepts, which can further extract scattering concepts. 

\textbf{Conceptualized Phrase Mining.} It is difficult to utilize rule-based approaches to mine long-tail instances and concepts. We propose the use of phrase mining methods to extract such instances and concepts.  The motivation is that \emph{there are numerous none-entity phrases in search queries, and such phrases play a vital role in understanding the query}.  Specifically, we firstly filter stop words and then employ an off-the-shelf phrase mining tool, AutoPhrase\footnote{\url{https://github.com/shangjingbo1226/AutoPhrase}},   to perform phrase mining on the corpora unsupervisedly. However, the original phrase mining approach experiences several issues in the real data set.   First, the results of the shallow syntax analysis of part-of-speech (POS) are not applicable to all areas, such as the medical domain.  Second, there is considerable wrong segmentation of words for Chinese. To address those issues, we use the existing domain knowledge graph to generate weak-supervised samples to re-train the POS and use the results of the POS tagging as well as  Chinese BERT-base embeddings as segmentation features for AutoPhrase. As AutoPhrase relies on distance supervised data, we leverage the data of the existing domain knowledge graph along with domain rules to generate positive and negative data. After running the Autophrase, we also propose a new empirical  score function with length constraint to generate high-quality phrases:
\begin{equation}
f(p)=\frac{1}{n}  \sum_{i=1}^{n} p_{score}+\log \left( p_{len}\right)
\end{equation}

Where $p_{score}$ and $p_{len}$ refer to the score and length of separated  phrase $p$, respectively.  Based on the mined phrases, we train a concept classification model based on Fasttext\footnote{\url{https://fasttext.cc/}} for specific concepts (200 concepts in the vertical domain). Based on the classification model, we link concept instances with specific concept labels (e.g., \emph{body integrity identity disorder} isA \emph{rare mental disorder}). We generate positive training samples of the concept classifier from those instances existing in the conceptual graph and create negative samples with domain rules (e.g., add negative character).     
 
\textbf{Self-Training with an Ensemble Consensus.}  Those unsupervised approaches are still limited in generalization. We further perform supervised learning. However, there is a lack of sufficient training samples.  Thus, we propose a novel low-resource tagging algorithm, namely self-training, with an ensemble consensus.  We leverage a few instances/concepts generated from the unsupervised approaches described in the section \ref{link} and \ref{sec-long-tail} as seeds (there may be only a few samples for some concepts in unique domains such as medical domain). Then, we train a CRF model\footnote{\url{https://taku910.github.io/crfpp/}} with seeds. We utilize the CRF model to generate a large amount of weak-supervised pseudo-sample from a subset of unlabelled data. Then, we train a BERT tagging model with permutations based on that pseudo-sample.  Thereafter, we generate pseudo-samples once more. Different from the first stage, we combine the CRF, BERT tagging, and a domain dictionary model (maximum forward match) with an ensemble consensus to generate more confident training samples. In other words, the prediction of three different models agrees with each other; thus, such pseudo-label may be more confident.  Next, we train a BERT tagging model once more. After several iterations, we can finally obtain a high-performance sequence tagging model with only a few initial training seeds. 

\begin{algorithm}[th]
\begin{algorithmic}[1]
\caption{Self-training with an ensemble consensus} 
\Require Only few initial training samples  $L$, unlabeled data $U$,  domain dictionary $D$, randomly initialized parameter of BERT tagging model $\theta$, perturbation $\delta$
\State Train the CRF with $L$, build dictionary tagging model with $D$
\State Random sample subset $\mathcal{U}$  from $U$ and generate pseudo-sample $\mathcal{U}^l$ with CRF
\For{\emph{iter} iterations}

\For{minibatch index $b=1: \frac{\left|\mathcal{U}^{l}\right|}{\text { batchsize }}$}
\State Sample a subset $\mathcal{B}_{b}^{\prime}$ with batch size  $b$ from $\mathcal{U}^{l}$
\State $\theta \leftarrow \theta+\delta$
\State $\theta \leftarrow \theta-\alpha \nabla_{\theta} \frac{1}{\left|\mathcal{B}_{b}^{\prime}\right|} \sum_{(x, y) \in \mathcal{B}_{b}^{\prime}} C E\left(f_{\theta}(x), y\right)$
\EndFor
\State Random sample subset $\mathcal{U}$  from $U$

\State Generate pseudo-sample $\mathcal{U}^{ensemble}$ with voting  of dictionary tagging model, CRF model, and BERT tagging model
\State $\mathcal{U}^l \leftarrow  \mathcal{U}^l+\mathcal{U}^{ensemble}$
\EndFor
\Return BERT sequence tagging model $\theta$ 
\label{tag} 
\end{algorithmic}
\end{algorithm}

The above approaches for concept mining are complementary to each other. Our experience shows that bootstrapping with alignment consensus can extract head concepts with high accuracy. We also observe that for those long-tail concepts which occupies a large proportion of the search log, conceptualized phrase mining is advantageous for extracting those low-frequent terms, while sequence tagging can better extract concepts from the short text when they have a clear boundary with surrounding non-concept words. 

\subsection{Taxonomy Evolution\label{dynamic}}
 
 Existing taxonomies are mostly constructed by human experts or in a crowdsourcing manner. As the web content and human knowledge are constantly growing over time, people need to update existing taxonomy and also include new emerging concepts, and it is intuitive to incorporate the temporal evolution into the taxonomies. The key to handling the problem of taxonomy evolution is to properly model the probability of assigning a concept to a parent concept in the taxonomy tree, i.e., estimating the concept distribution, which can evolve.

\begin{figure}
\centering
\includegraphics [width=0.5\textwidth]{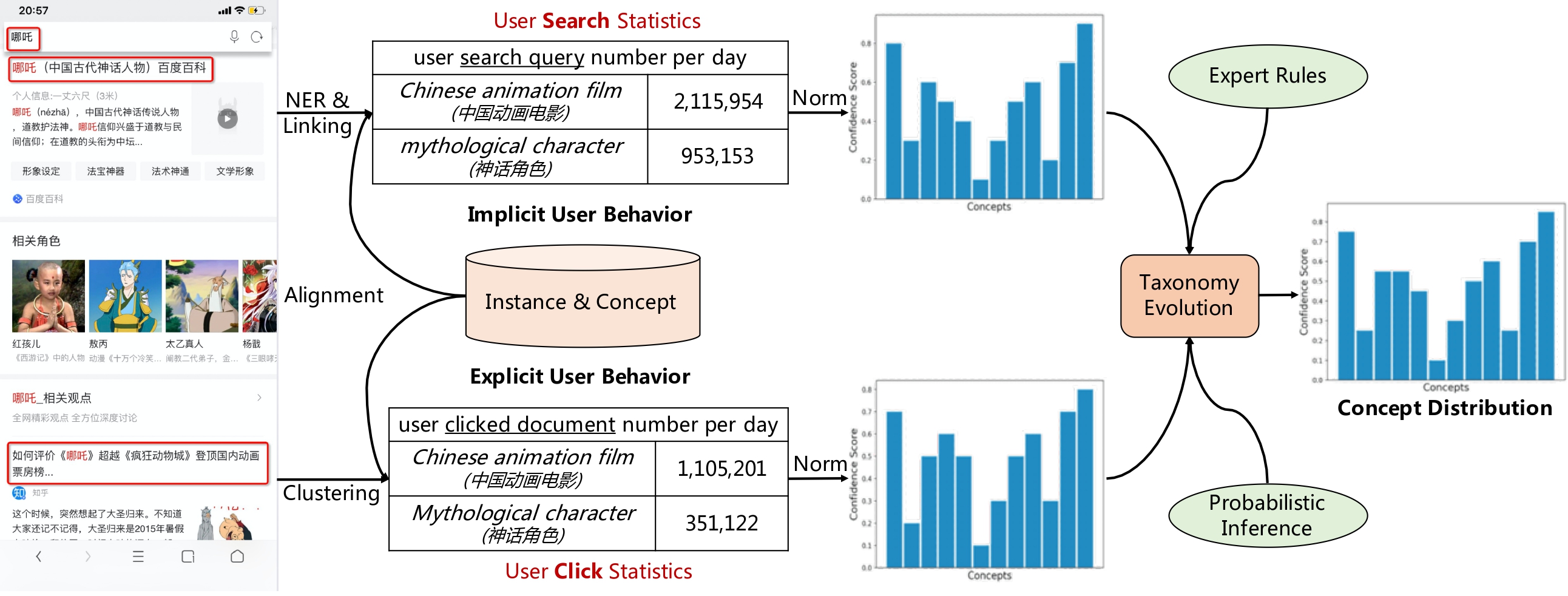}
\caption{Example of taxonomy evolution.}
\label{arc1_3}
\end{figure}

\textbf{Concept distribution estimation based on implicit and explicit user behavior.} Different from the previous approach, such as Probase \cite{wu2012probase}, we leverage the user behaviors to estimate the confidence score of concepts given instances. As the user search and click are important behaviors that reveal the user interests in search engines, it is intuitive to estimate confidence score with statistics of implicit user behaviors such as searching and explicit user behaviors such as clicking on the documents.  As Figure \ref{arc1_3} shows,  we first align parts of concepts with a predefined synonym dictionary as part of the concepts and instance are from different sources, which result in redundancy. Then, we calculate the confidence scores via the heat of entities. We estimate the heat of entities via user searching per day, and the details are included in the appendix. Specifically, given instances with concepts $c \in C$, we run named entity recognition for each $c$ (each concept only remains one most important entity, e.g., largest TF-IDF score)? And obtain a heat score. Then we calculate the implicit concept distribution based on user search behaviors as follows:
\begin{equation}
s_{implicit}=\frac{c_{heat}}{\sum_{c \in C} c_{heat}}
\end{equation}
where $c_{heat}$ means the heat of the entity in concept $c$ and $C$ is the concept set that  $c$ belongs to. Then, we leverage user clicking behaviors to obtain concept distributions explicitly.  Specifically, we first collect one month's instances (queries) clicked tag pairs and then aggregate the of queries with the same clicked concept tag?\footnote{The search logs contain the history of user clicked concept tags.}. In other words, we estimate the number of queries for each concept tag in the last month. For example, if users clicked more "Chinese animation film (\begin{CJK}{UTF8}{gbsn}中国动画电影\end{CJK})" when searching "Nezha (\begin{CJK}{UTF8}{gbsn}哪吒\end{CJK})", then the instance "Nezha (\begin{CJK}{UTF8}{gbsn}哪吒\end{CJK})" will be more confident with concept "Chinese animation film (\begin{CJK}{UTF8}{gbsn}中国动画电影\end{CJK})". Thus, we can estimate the concept of confidence distribution based on user clicking behavior. Finally, we combine two different granularity confidence scores as follows:
\begin{equation}
s= \left\{ \begin{array}{l}{\sum_{i}^{N}\left(s_{\text {implicit}}+s_{\text {explicit}}\right) \text {if } s \notin M} \\ {\sum_{i}^{N}\left(1.5 \log \left(s_{\text {implicit}}\right)+\log \left(\mathrm{s}_{\text {explicit}}\right) \text {if } s \in M\right.}\end{array}\right.
\end{equation}

Where $s_{implicit}$ is the confidence score based on user searching behavior, $s_{explicit}$ is the confidence score based on the user clicking behavior, and $M$ is the specific subset (e.g., medical and education domain) of the concept set. The specific subset is required here because some domains have special user behaviors, for example, no clicking or lots of misleading clicking. We also remove concepts such as "stop words" and "common words" to prevent unnecessary noise. Meanwhile, we re-weight domain words and specific types of concepts that are unambiguous by $s = 19.5log (c_{heat})s$,  e.g., "\emph{Pauli incompatible}" isA "\emph{chemical term}" and "\emph{Andy Lau}" isA "\emph{person}."  

After that, we can build the instance--concept taxonomy in an evolving way. Then we leverage expert rules to define the taxonomy between \textbf{level1} and \textbf{level2}. We further determine the relation between \textbf{level2} and \textbf{level3} via probabilistic inference. Specifically, suppose there is a concept  $c$, a higher level concept $p$, $n_c$ instance  related to concept $c$, and $n_p^c$ instance belonging to concept $p$. Then, we estimate $p(\mathbf{p} | \mathbf{c}) \text { by } p(\mathbf{p} | \mathbf{c})=n_{\mathbf{p}}^{\mathbf{c}} / n^{\mathbf{c}}$. We identify the isA relation between $c$ and $p$ if $p(\mathbf{p} | \mathbf{c})>\delta_{t}$ .  

\section{Deployment\label{deploy}}
We propose three methods for deploying the conceptual graph at Alibaba, namely \textbf{text rewriting}, \textbf{concept embedding}, and \textbf{conceptualized pretraining}. 

\textbf{Text Rewriting.} For each text instance $s$, we extract the concept  $c$ conveyed in the text and rewrite the text by concatenating each instance with $s$. The rewritten text format is $s$ $c$. We use text rewriting for information retrieval. Note that text rewriting is easy to apply to other classification or sequence labeling tasks. 

\textbf{Concept Embedding.}  
Following \cite{chen2019deep}, we utilize a two-tower neural network with concept attention and self-attention for learning concept embedding. Then we concatenate the concept embedding with text embedding for sub-tasks. 


\textbf{Conceptualized Pretraining.} As pretraining is quite powerful, the conceptual graph can be utilized during the pretraining stage to inject knowledge explicitly.  Specifically, we utilize instances and concept masking strategies with an auxiliary concept prediction loss to integrate conceptual knowledge. 
  
\section{Evaluation}
In this section, we first introduce a new dataset for the problem of concept mining from the search logs and compare the proposed approach with various baseline methods. Then, we evaluate the accuracy of the taxonomy evolution. We also evaluate different methods to apply conceptual graphs in applications, including intent classification and named entity recognition. Finally, we perform large-scale online A/B testing to show that our approach significantly improves the performance of the semantic search. 

\subsection{AliCG Construction Evaluation}

\subsubsection{Evaluation of Fine-grained Concept Acquisition}\  \\
We evaluate our approach on two datasets. The first is the user-centered concept mining (UCCM\footnote{\url{https://github.com/BangLiu/ConcepT}}) dataset \cite{liu2019user}, which is sampled from the queries and query logs of Tencent QQ Browser. The second is the Alibaba Conceptual Graph (ACG) dataset\footnote{\url{https://github.com/alibaba-research/ConceptGraph}}. We have created a large-scale dataset containing 490,000 instances with the fine-grained concept, which is sampled from the search logs of Alibaba UC Browser from November 11, 2018, to July 1, 2019.

 We compare our approach with the following baseline methods, and the variants of our approach: \textbf{TextRank} \cite{mihalcea2004textrank} is a  classical graph-based ranking model for keyword extraction;  \textbf{THUCKE} \cite{liu2011automatic} is a method that considers keyphrase extraction as a problem of translation and learns translation probabilities between the words in the input text and the words in keyphrases; \textbf{AutoPhrase} \cite{shang2018automated} is a  quality phrase mining algorithm that extracts quality phrases based on a knowledge base and POS-guided segmentation;  \textbf{ConcepT} \cite{liu2019user} is a state-of-the-art concept mining approach that utilizes bootstrapping, query-title alignment, and sequence tagging methods to exact concepts from search queries;   \textbf{w/o BA)} is the approach that extracts concepts from search logs without \textbf{\emph{Bootstrapping with the Alignment consensus}}.

\begin{figure*} \centering
\subfigure[Wuhan] { \label{aa}
  \includegraphics[width=0.23\textwidth]{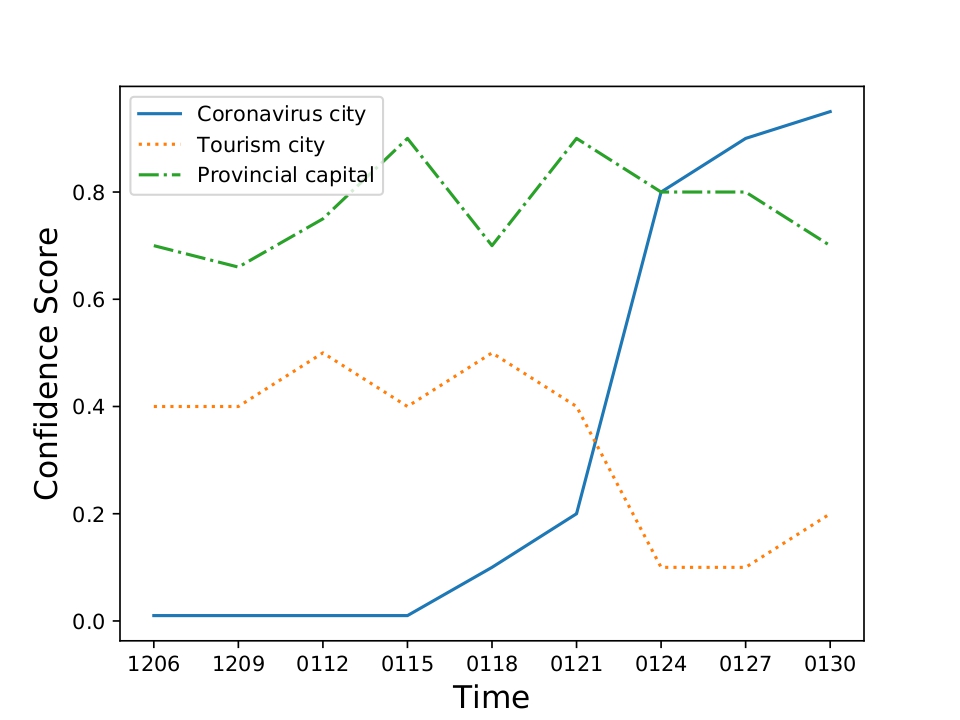}
}
\subfigure[Shandong carrier] { \label{bb}
\includegraphics[width=0.23\textwidth]{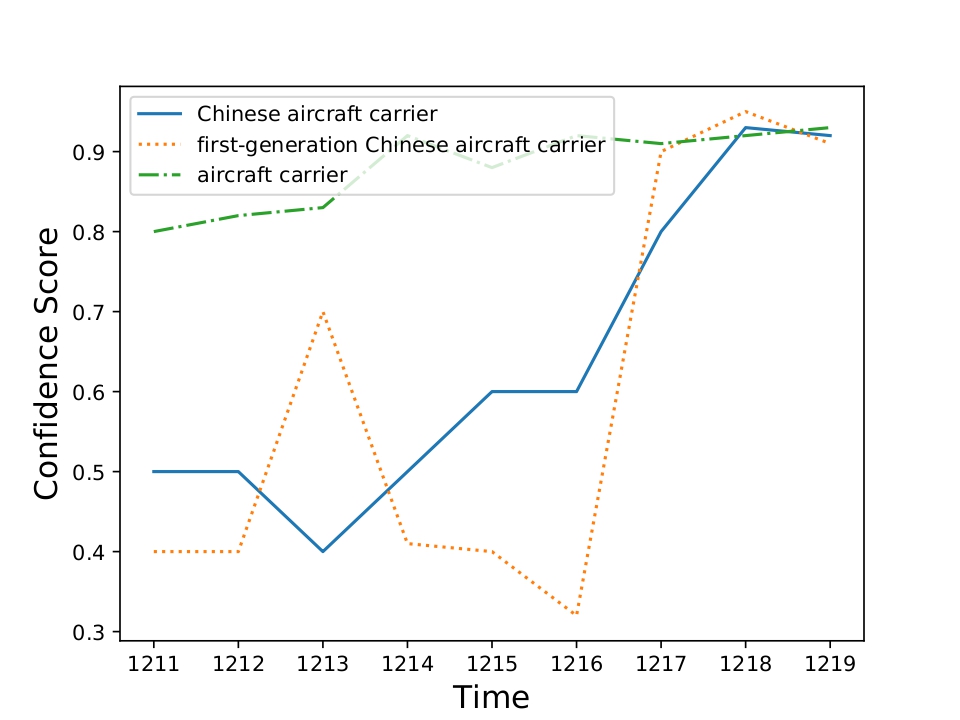}
}
\subfigure[Nezha] { \label{cc}
\includegraphics[width=0.23\textwidth]{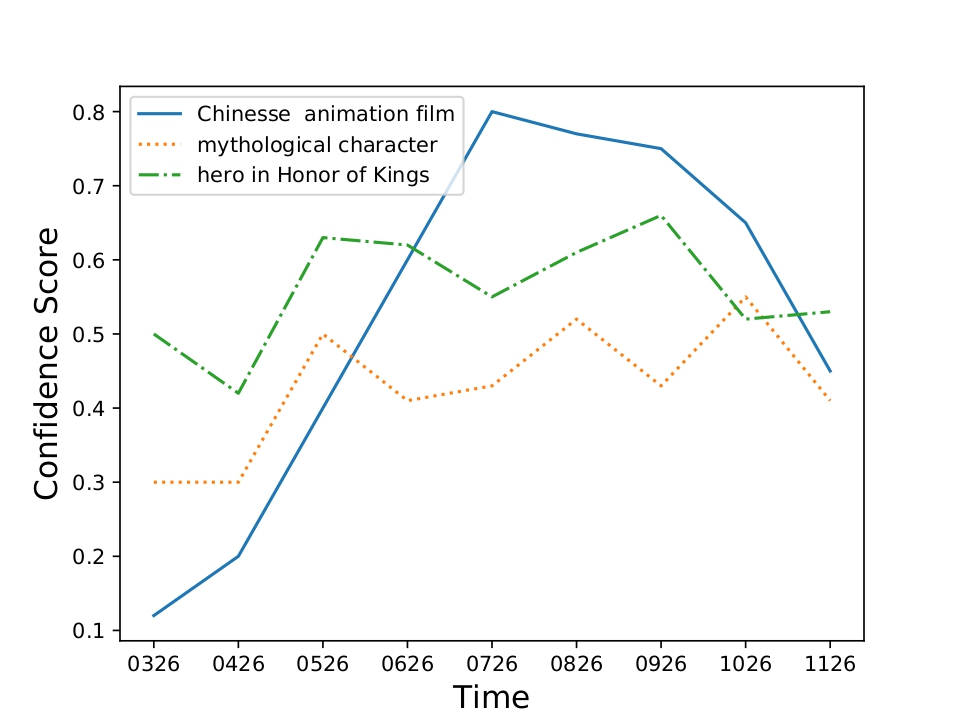}
}
\subfigure[NCP] { \label{dd}{}
\includegraphics[width=0.23\textwidth]{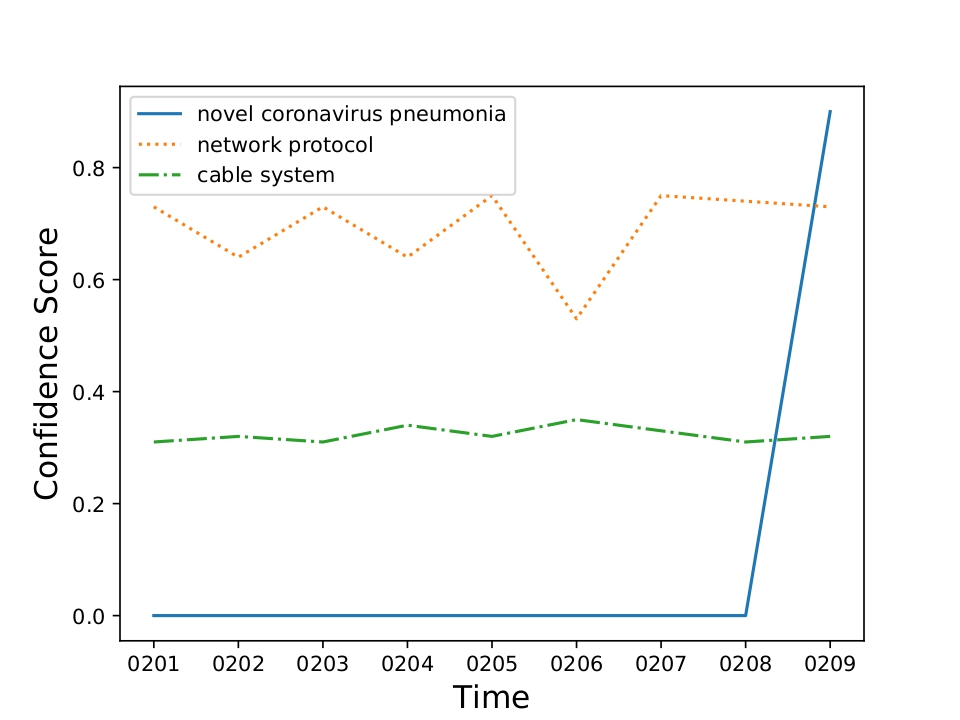}
}
\caption{Concept evolution visualization.}
\label{vis}
\end{figure*}

\begin{table}
     \caption{Evaluation results of  fine-grained Aquizition.}
    \label{mining}
    \small
    \centering
\begin{tabular}{c | c c | c c}
\toprule
Dataset & \multicolumn{2}{c|}{UCCM} & \multicolumn{2}{c}{ACG} \\
\midrule
Method         & Exact Match & F1 Score & Exact Match & F1 Score \\
\midrule
TextRank     & 0.1941 & 0.7356 & 0.0025 & 0.1839 \\ 
THUCKE         & 0.1909 & 0.7107 & 0.0325 & 0.2839 \\
AutoPhrase     & 0.0725 & 0.4839 & 0.0325 & 0.2839 \\
ConcepT     & 0.8121 & 0.9541 & 0.7725 & 0.7839 \\
\hline
AliCG & \textbf{0.9122} & \textbf{0.9812}& \textbf{0.9215} &\textbf{0.9898}\\
\hline

w/o BA & 0.8785 & 0.9635 &  0.8852 & 0.9543 \\

\bottomrule
\end{tabular}
\end{table}
 
From Table \ref{mining}, we observe that our method achieves the best \emph{Exact Match (EM)} and \emph{F1 scores}. This is because that guiding and aligning consensus help us build a collection of high-quality instances and concepts in an unsupervised manner.  The TextRank, THUCKE, and AutoPhrase methods do not provide satisfactory performance because they are more suitable for extracting keywords or phrases from long documents or corpora. Our method exhibits better performance compared with ConcepT mainly because our method can extract more accurate concepts with consistency in query-title pairs.  

\begin{table}
    \caption{Concept mining examples.}
    \label{example}
    \centering
    \small
\begin{tabular}{c | c}
 
\toprule

\textbf{Instance} & \tabincell{l}{People with \emph{rare mental disorder} such as \textbf{aboulomania} \\ and \textbf{body integrity identity disorder} appear generally \\ normal but when faced with … } \\

\midrule
\hline
\textbf{Model} & Instance-Concept \\
\hline

AutoPhrase     & No results \\ 
ConceptT     & (\emph{aboulomania} \textbf{isA} \emph{rare mental disorder}) \\
\hline
AliCG         & \tabincell{c}{(\emph{body integrity identity disorder} \textbf{isA} \emph{rare mental disorder}) \\ (\emph{aboulomania} \textbf{isA} \emph{rare mental disorder})} \\

\bottomrule
\end{tabular}
\end{table}

\begin{table}
    \caption{Evaluation results of long-tail concept mining on ACG (long-tail) dataset.}
    \label{mining2}
    \small
    \centering
\begin{tabular}{c | c c}
\toprule
Method & Exact Match & F1 Score \\
\midrule
TextRank     & 0.0941 & 0.1356 \\ 
THUCKE         & 0.1209 & 0.2107 \\
AutoPhrase     & 0.1725 & 0.2839 \\
ConcepT     & 0.3121 & 0.2541 \\
\midrule
AliCG & \textbf{0.6122} & \textbf{0.8812} \\
\hline
w/o CP & 0.5522 & 0.7612 \\
w/o SE & 0.5319 & 0.7322 \\

\bottomrule
\end{tabular}
\end{table}
 
\subsubsection{Evaluation  of Long-tail Concept Mining} \ \\
As there are no public datasets available for long-tail concept mining, we leverage our ACG dataset and randomly choose 100 long-tail concepts with corresponding texts to build the dataset ACG (long-tail). We evaluate our long-tail concept mining methods on that dataset compared with the same baselines in fine-grained concept mining and the variants of our approach. \textbf{w/o CP} is the approach that extracts concepts without \textbf{\emph{Conceptualized Phrase mining}};   \textbf{w/o SE} is the approach that extracts concepts without \textbf{\emph{Self-training and Ensemble consensus}}. 
 
From  Table \ref{mining2}, we observe that our method achieves the best EM and F1 scores. Firstly, the conceptualized phrase mining method extracts numerous long-tail instances and concepts. Secondly, self-training sequence tags with an ensemble consensus can identify scattered concepts from the search logs; that is, the conceptual boundaries in the text are not clear. The TextRank, THUCKE, and AutoPhrase methods fail to perform well in ACG (long-tail), which indicates the difficulty of long-tail concept extraction.   Our method exhibits significant improvement compared with ConcepT mainly because our method can extract long-tail concepts and achieve excellent performance in the sequence labeling with only a small number of samples. The comparison shows that our method outperforms its variants and proves the effectiveness of combining different strategies in our system.  We also show examples of concept mining in Table \ref{example}, which illustrates that our approach is more capable of extracting fine-grained concepts.

\subsubsection{Evaluation of Taxonomy  Evolution} \ \\ 
We randomly extract 1,000 instances from the taxonomy. The experiments focus on evaluating the relation between concept--instance pairs because these relations are critical for semantic search. We check whether the isA relation between each concept and its instance is correct. We ask three human judges to evaluate the relations. We record the number of correct and incorrect instances for each concept. Table \ref{taxonomy} shows the results of the evaluation. The average number of instances per concept is 5.44, and the most significant concept contains 599 instances. Note that the size of the taxonomy is increasing and updated with daily user query logs. The accuracy of the isA relation between concept--instance pairs is 98.59\%. We also notice that implicit and explicit user behaviors (\textbf{w/o implicit} or \textbf{w/o explicit}) have a significant loss in performance, which indicates that user behavior plays an important role in concept distribution estimation. 

\begin{table}
    \caption{Evaluation results of taxonomy evolution.}
    \label{taxonomy}
    \small
    \centering
\begin{tabular}{c c}
 
\toprule
Metrics / Statistics & Value \\ 
\midrule
Mean \#Instances per Concept & 5.44 \\
Max \#Instances per Concept &  599 \\
isA Relationship Accuracy & \textbf{98.59} \\
\hline
w/o implicit &  95.23 \\
w/o explicit &  93.78 \\
w/o both &  89.12 \\
\bottomrule
\end{tabular}
\end{table}
 
We visualize the concept trending over time to evaluate the dynamic of AliCG. From Figure \ref{vis} we observe that: 1) 
our approach can obtain a dynamic and fine-grained concept distribution over time. Figure \ref{aa} shows that the concept of "Wuhan (\begin{CJK}{UTF8}{gbsn}武汉\end{CJK})" is more confident with "coronavirus city  (\begin{CJK}{UTF8}{gbsn}冠状病毒城市\end{CJK})" after January 22th, while it is confident with "tourism city (\begin{CJK}{UTF8}{gbsn}旅游城市)\end{CJK}" before January 22th. Note that there was a severe coronavirus outbreak in Wuhan, and lots of relevant news published on the web after January 22th.  2)  Our approach can also obtain false confident scores due to the noise in user behaviors. Figure \ref{bb} indicates that "Shandong carrier (\begin{CJK}{UTF8}{gbsn}山东号\end{CJK}") has a low confidence score in the day before December 17. Such circumstance is caused by some fake or  clickbait news which misleads user behaviors, and it results in the wrong confidence score of "first-generation Chinese aircraft carrier (\begin{CJK}{UTF8}{gbsn}中国第一代航母\end{CJK})." However, our approach can self-repair and get the correct confident score after that day.  3) Our approach can reveal the concept distribution trend over a long period of time. From Figure \ref{cc}, we observe the confidence score of "Nezha (\begin{CJK}{UTF8}{gbsn}哪吒\end{CJK})" is becoming higher with "Chinese animation film (\begin{CJK}{UTF8}{gbsn}中国动漫电影\end{CJK})," but it started to slightly decreasing after July, which reveals the attenuation of user interest over time.  4) Our approach can extract emerging concepts. We observe from \ref{dd} that our approach extract the emerging concept "novel coronavirus pneumonia (\begin{CJK}{UTF8}{gbsn}新冠肺炎\end{CJK})" on February 8. Before that, the concept of "NCP" is mostly confident with "network protocol (\begin{CJK}{UTF8}{gbsn}网络协议\end{CJK})" or "cable system (\begin{CJK}{UTF8}{gbsn}电话系统\end{CJK})." 
 
 \subsection{AliCG Deployment Evaluation}
 AliCG is currently deployed at Alibaba to support a variety of business scenarios, including the product Alibaba UC Browser.  The current system can extract approximately  20,000 concepts every day and serve more than 300 million daily active users.  AliCG has been applied to dozens of applications, such as intent classification, named entity recognition, information retrieval, entity recommendation, and so on.  We will introduce the evaluation of those scenarios to demonstrate the efficacy of AliCG.

\begin{figure} \centering
\subfigure[Intent Classification (Medical)] { \label{intent}
  \includegraphics[width=0.22\textwidth]{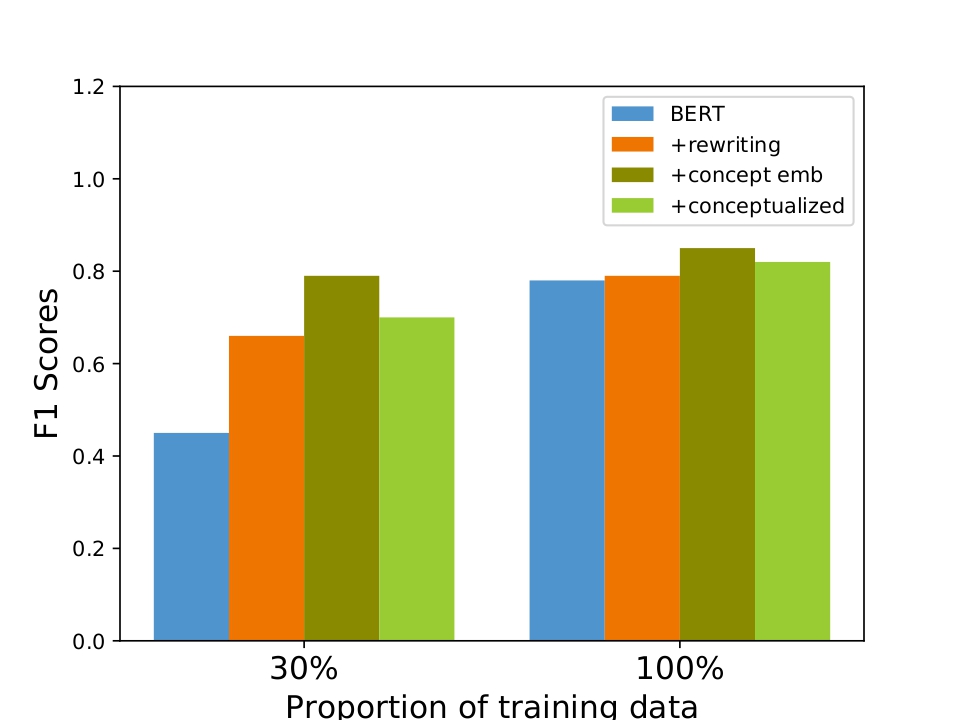}
}
\subfigure[Named entity recognition (Medical)] { \label{ner}
\includegraphics[width=0.22\textwidth]{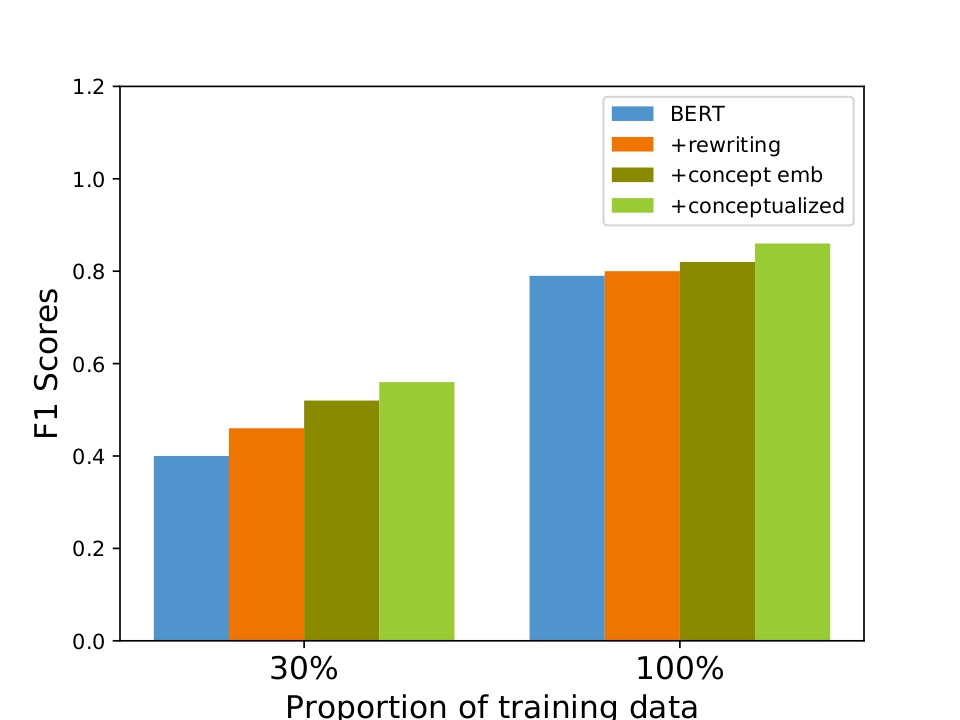}
}
 \subfigure[Intent Classification (General)] { \label{intent2}
  \includegraphics[width=0.22\textwidth]{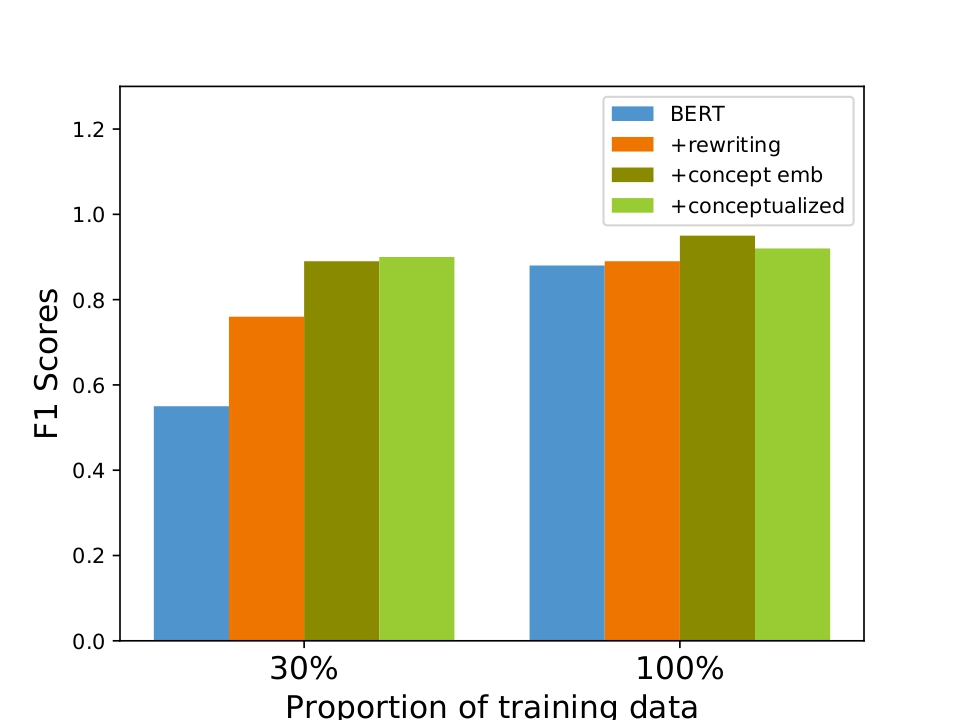}
}
\subfigure[Named entity recognition (General)] { \label{ner2}
\includegraphics[width=0.22\textwidth]{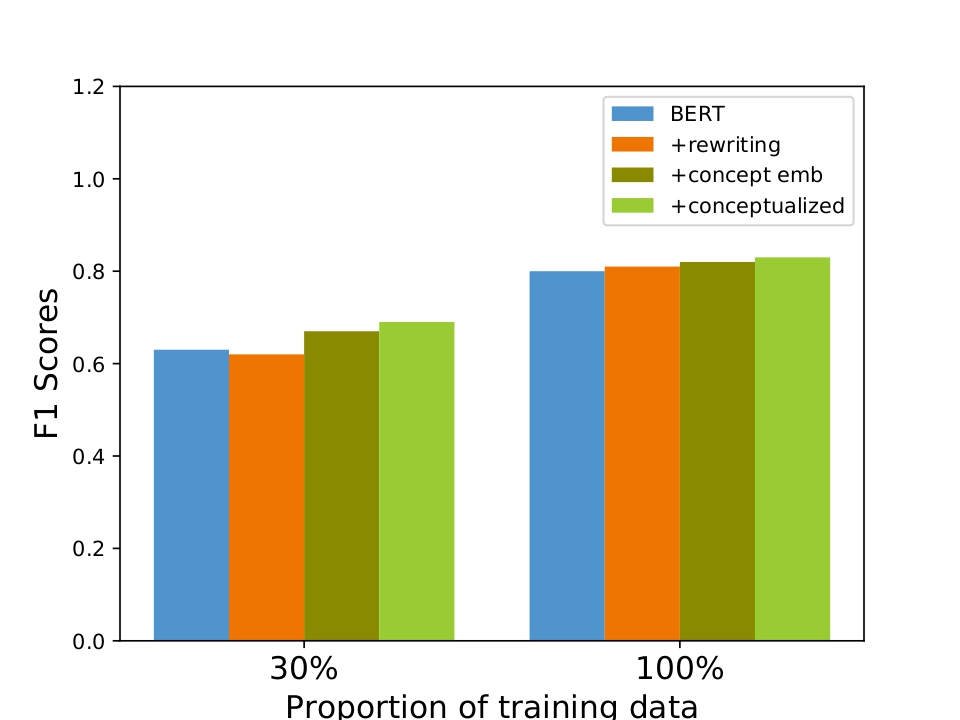}
}
\caption{Evaluation results of applications.}
\label{exp_knowledge}
\end{figure}

\subsubsection{Intent Classification and Named Entity Recognition} \ \\
We apply AliCG to two tasks to evaluate the different deployment methods in both specific domains (medical domain\footnote{\url{https://github.com/alibaba-research/ChineseBLUE}}) and general domain (search logs). \textbf{+rewriting} is the method with text rewriting;  \textbf{+concept emb} is the method with concept embedding;  \textbf{+conceptualized} is the method with conceptualized pretraining. From Figure \ref{exp_knowledge}, we observe: 1)  All knowledge-enhanced approaches achieve better performance compared with the baseline BERT, which demonstrates that conceptual knowledge is advantageous for sub-tasks. 2) The performance gains more improvement in the medical domain compared with the general domain. We speculate that a specific domain has different data distribution, which indirectly improves the impact of knowledge injection. 3) With only 30\% of the training data, the knowledge-enhanced model can achieve comparable performance compared with baselines with 100\% training data. This result reveals that knowledge-enhanced approaches are beneficial in a low-resource setting, which is quite useful as specific domains usually are data-hungry. 4) In the medical domain,  the method with concept embedding achieves the best performance in intent classification, while the method with conceptualized pretraining is 2\% higher than the method with concept embedding in named entity recognition. We argue that this is because named entity recognition is a sequence tagging task, and conceptualized pretraining injects prior knowledge via masking strategies, which boosts the performance. However, intent classification is a classification approach, and there are distinct features that affect the final prediction. In addition, the attention mechanism may contribute to the performance.

\begin{table}
    \caption{Online evaluation results.}
    \label{online}
    \small
    \centering
\begin{tabular}{c c | c c}
\toprule
\multicolumn{2}{c|}{Information Retrieval} & \multicolumn{2}{c}{Entity Recommendation} \\ 
\midrule
Search Engine & Relevant Precision & Metrics & Percentage Lift \\ 
\midrule
Baidu & 72.5 & CTR & \textbf{5.1}\% \\
Baidu+AliCG & \textbf{84.1} & PV & \textbf{5.5}\% \\
Shenma & 71.1 & UD & \textbf{0.92}\% \\
Shenma+AliCG & \textbf{83.1} & IE & \textbf{7.01}\% \\
\bottomrule
\end{tabular}
\end{table}

\subsubsection{Online Evaluation of  Information Retrieval} \ \\
We evaluate how concept mining can help improve information retrieval through text rewriting. We create an evaluation dataset containing 200 queries from the Alibaba UC browser. For the original query, we collect the top 10 search results returned by the Baidu\footnote{\url{https://www.baidu.com/}} and Shenma\footnote{\url{https://m.sm.cn/}} search engines, which are the first and second largest search engines in China, respectively. We first replace the query with $k$ different instances and collect the top search results for each rewritten query from Baidu and Shenma. We then merge and retain 10 of them as search results after query rewriting. We ask three human judges to assess the relevance of the results and record the majority vote of the human judges, which is "relevant" or "irrelevant," and calculate the percentage of the relevance of the original query and rewritten query. As shown in Table \ref{online}, for the query after rewriting using our strategy, the percentage of top 10 relevant results increases from \textbf{72.5\%} to \textbf{84.1\%} in Baidu and from \textbf{71.1\%} to \textbf{83.1\%} in Shenma respectively. The reason is that conceptual knowledge helps understand the intent of user queries, which can provide more relevant and clear keywords for search engines. As a result, information retrieval results in a better match for users’ intent.

\subsubsection{Online A/B Testing of Entity  Recommendation} \ \\ We perform a large-scale online A/B testing to show how conceptual knowledge helps in improving the performance of entity recommendations (see appendix for details) in real-world applications \cite{zhang2021cause}.  We divide users into buckets for the online A/B test. We observe and record the activities of each bucket for seven days and select two buckets with highly similar activities. We utilize the same text rewriting and conceptualized pretraining strategies in section \ref{deploy}  for entity recommendation.  The recommendations are obtained without the conceptual graph for one bucket and with the conceptual graph for the other bucket. The page view (PV) and click-through-rate (CTR) are the two most critical metrics in real-world applications because they show how many content users there are and how much time they spend on an application. We also report the  Impression Efficiency (IE) and Users Duration (UD).  As shown in Table \ref{online}, we observe a statistically significant increase in  CTR (\textbf{5.1\%}),  PV (\textbf{5.5\%}), UD (\textbf{0.92\%}) and IE (\textbf{7.01\%}).  These observations prove that the conceptual graph for entity recommendation considerably benefits the understanding of queries and helps match users with their potential interested entities. 

\section{Related Work} 
\textbf{Fine-graied Concept Acquisition.} 
 Conventional concept mining methods are closely related to noun phrase segmentation and named entity recognition \cite{DBLP:journals/corr/abs-2009-06207}. They either use heuristic methods to extract typed entities or treat the problem as sequence labeling and use large-scale labeled training data to train LSTM-CRF. Another area of focus is term and keyword extraction. They extract noun phrases based on statistical appearance and co-occurrence signals \cite{frantzi2000automatic} or text features \cite{medelyan2006thesaurus}. The latest methods for concept mining rely on phrase quality. \cite{shang2018automated} adaptively identified concepts based on their quality.   
 More recently, \cite{liu2019user}  propose an approach to discover user-centered concepts via bootstrapping, query-title alignment, and sequence labeling. Different from their approach, our method is advantageous to extract long-tail concepts and also have the ability to evolve the taxonomy.  
 
\textbf{Long-tail Concept Mining.}  Conventional approaches usually fail to extract long-tail concepts.  Some existing approaches leverage unsupervised approaches to extract long-tail concepts as phrase mining \cite{shang2018automated}.  However, those approaches can neither leverage the context nor handle scattered concepts. Besides, some studies  \cite{liu2019user} treat concept mining as sequence tagging. Nevertheless, the performance of those long-tail concepts degrades significantly.  Several low-resource sequence tagging approaches that utilize transfer learning \cite{cao2019low}, or meta-learning \cite{hou2019few}   have been proposed. However, there are few Chinese resources available, and the meta-learning approach still suffers from low accuracy, which is not satisfactory for real applications. 

\textbf{Taxonomy Evolution.} Automatic taxonomy construction is a long-standing task in the literature. Most existing approaches focus on building the entire taxonomy by first extracting hypernym-hyponym pairs and then organizing all hypernymy relations into a tree or DAG structure.  In many real-world applications, some existing taxonomies may have already been laboriously curated by experts \cite{lipscomb2000medical} or via crowdsourcing \cite{meng2015crowdtc}, and are deployed in online systems. Instead of constructing the entire taxonomy from scratch, these applications demand the feature of expanding an existing taxonomy dynamically. There exist some studies on expanding WordNet with named entities from Wikipedia or domain-specific concepts from different corpora \cite{jurgens2015reserating}.   One major limitation of these approaches is that they cannot update the taxonomy dynamically, which is necessary for semantic search.  Probase \cite{wu2012probase} proposes two probability scores, namely \emph{plausibility}  and \emph{typicality}, for the conceptual graph.  However,  it is computationally expensive and time-consuming to update such a big conceptual graph. 

\section{Conclusion}

We describe the implementation and deployment of AliCG at Alibaba, which is designed to improve the performance of the semantic search. The system extracts fine-grained concepts from the search logs, and the extracted concepts can be updated based on user behaviors. We have deployed the conceptual graph at Alibaba UC Browser with more than 300 million daily active users. Extensive experimental results show that the system can accurately extract concepts and boost the performance of intent classification and named entity recognition, and Online A/B testing results further demonstrate the efficacy of our approach. 


\section{Acknowledgements}


This work is supported by the National Natural Science Foundation of
China (Nos. 91846204, 52007173 and U19B2042), National Key R\&D Program of China (No. SQ2018YFC000004), Zhejiang Provincial Natural Science Foundation of China (No. LQ20E070002), and Zhejiang Lab's Talent Fund for Young Professionals (No. 2020KB0AA01).



\bibliography{KDD2020}
 
\bibliographystyle{ACM-Reference-Format}
\clearpage

\appendix 

\section{INFORMATION FOR REPRODUCIBILITY}

\subsection{System Implementation and Deployment}

We implement and deploy the AliCG system in Alibaba UC Browser. The fine-grained concept acquisition module, long-tail concept mining module,  taxonomy evolution module are implemented in Python 3.6 and run as offline components.  The concept embedding and conceptualized pretraining module (inference part) are implemented in C++, and they run as an online service. We utilize high-performance memory storage Tair (a Redis-like storage system) in Alibaba for online data storage. In our system, each component works as a service and is deployed on Alibaba MaxCompute\footnote{\url{https://www.alibabacloud.com/}}. Alibaba MaxCompute is a big high-performance data processing framework. It provides fast and fully managed petabyte-scale data warehouse solutions and allows for the analysis and processing of massive amounts of data economically and efficiently.  The online service runs on 200 dockers. Each docker is configured with six 2.5 GHz Intel Xeon Gold 6133 CPU cores and 64 GB memory. Offline fine-grained concept acquisition,  long-tail concept mining, and taxonomy evolution are run on 60 dockers with the same configuration. We set $\delta_t = 0.3$, $\alpha = 0.6$, $\beta = 0.8$, and $\delta = 2$ in our system.

\begin{algorithm}[th]
\begin{algorithmic}[1]
\caption{Offline instance/concept mining} 
\Require query and high-clicked documents set $ (q,D) \in T$.
\If{succeed} 
\State Perform  bootstrapping with alignment consensus on general domain
\State Perform conceptualized phrase tagging on specific domains
\State Perform self-training with an  ensemble consensus
\If{coverage}
\State Perform  taxonomy evolution
\State Perform refinement
\EndIf
\Else
\State Break
\EndIf
\label{alg3} 
\end{algorithmic}
\end{algorithm}

\begin{algorithm}[th]
\begin{algorithmic}[1]
\caption{Offline concept embedding}
\Require query and high-clicked documents set $ (q,D) \in T$.
\State Tag $T$ with all candidate concepts with maximum forward matching
\State Train concept embedding model 
\label{alg4} 
\end{algorithmic}
\end{algorithm}

\begin{algorithm}[th]
\begin{algorithmic}[1]
\caption{Offline conceptualized pretraining}
\label{algorithm: training}
\Require query and high-clicked documents set $ (q,D) \in T$,  conceptual graph $CG$ stored in key-value memory storage
\State Tag $T$ with all candidate concepts by maximum forward matching from $CG$

\State Duplicate and shuffle the corpus $T$ ten times and generate whole concept masking training samples with a masking rate of 15\%.

\State Initialize all parameters with BERT-wwm and perform further pretraining with $T$. 
\State Finetune sub-tasks with the  further pretrained model
\label{alg5} 
\end{algorithmic}
\end{algorithm}

\begin{algorithm}[th]
\begin{algorithmic}[1]
\caption{Online concept embedding inference}
\Require query set $Q$,  conceptual graph $CG$ stored in key-value memory storage, concept embedding  model $M$ in tf-serving cluster
\For{each query $q \in Q$}
\State Tag $q$ with all candidate concepts by maximum forward matching from $CG$
\State Call tf-serving model $M$ and get results
\EndFor
\label{alg6} 
\end{algorithmic}
\end{algorithm}

\begin{algorithm}[th]
\begin{algorithmic}[1]
\caption{Online concept pretraining inference}
\Require query set $Q$, conceptualised   model $M$ in tf-serving cluster
\For{each query $q \in Q$}
\State Call tf-serving model $M$ and get results
\EndFor
\label{alg7} 
\end{algorithmic}
\end{algorithm}

 \begin{figure}
\centering
\includegraphics [width=0.5\textwidth]{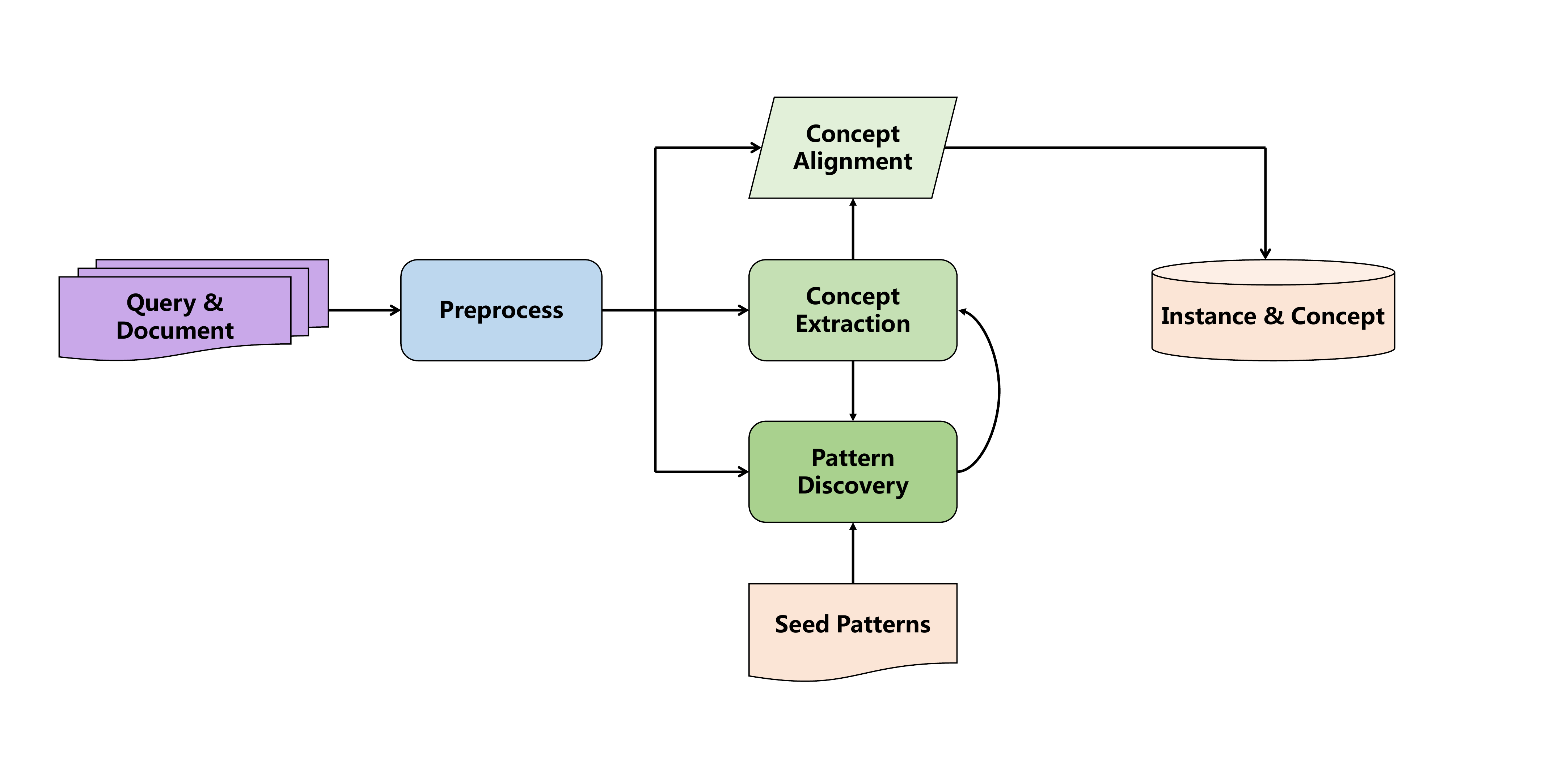}
\caption{Procedure of fine-grained concept acquisition.}
\label{arc1_1}
\end{figure}

\begin{figure}
\centering
\includegraphics [width=0.45\textwidth]{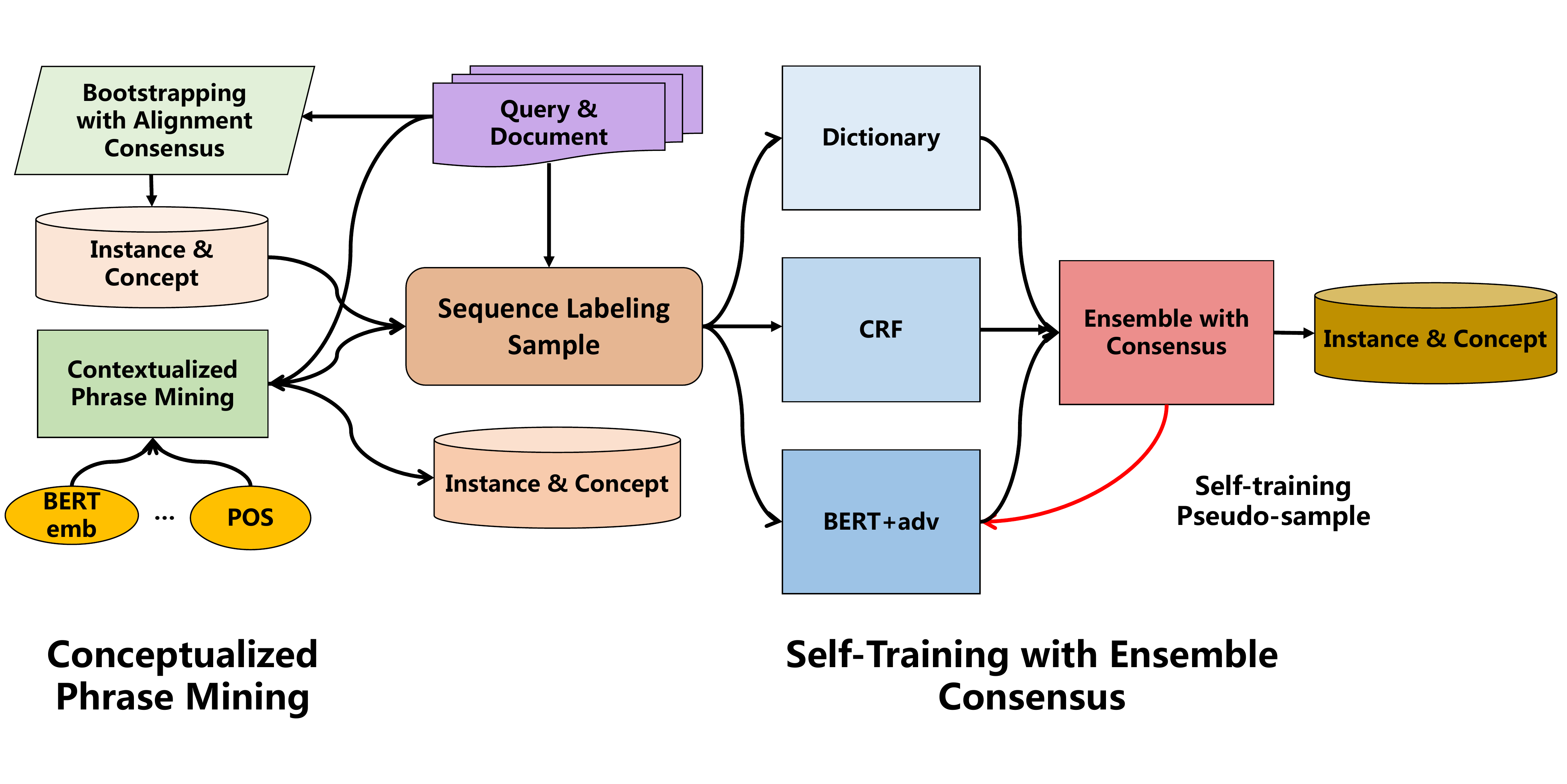}
\caption{Procedure of   long-tail concept mining.}
\label{arc1_2}
\end{figure}

Algorithms \ref{alg3}, \ref{alg4} and \ref{alg5} show the offline running processes of each component in AliCG. Algorithms \ref{alg6} and \ref{alg7} show the online running processes of concept embedding and conceptualized pretraining. We also detail the procedure of fine-grained concept acquisition and long-tail concept mining in Figure \ref{arc1_1} and Figure \ref{arc1_2}, respectively. Offline concept mining from the search logs is running every day. It extracts approximately 20,000 concepts from 30 million search logs, and approximately 5,000 of the extracted concepts are new. The offline taxonomy evolution is also daily running. 
The processing offline usually takes about two hours, and the processing time of online concept embedding is 10,0000 queries per second at most. It can perform concept embedding for approximately 800 million queries per day.

\subsection{Online Service Details}
We utilize a high-performance C++ server framework for online query embedding inference, which can handle more than 10,000 QPS. In addition to the regular offline update pipeline, we develop an emergency makeup pipeline, which has two functions. First, we mine concepts from heat articles (high clicked and page view articles) in real-time and put those heat concepts into the emergency makeup pipeline. In addition, we utilize this pipeline to fix a few bad cases. 

\subsection{Concept Reverse Index}
As some applications need to know the instances given concepts, for example, it is necessary to retrieve all the instances of the concept  "patriotic songs recommended by the Communist Youth League (\begin{CJK}{UTF8}{gbsn}共青团推荐爱国歌曲\end{CJK})" when user searching such a concept. Thus, we build a reverse concept--instance index based on HA3, which is a high-performance index builder at Alibaba. 


\subsection{Parameter Settings and Training Process}
Here, we introduce the features that we use for different components in our system and describe how we train each component.  For concept mining, we randomly sample 15,000 search logs in Alibaba UC Browser. We extract concepts from these search logs using the approaches introduced in Sec. \ref{link} and the results are manually checked by Alibaba product managers. The resulting dataset is used to train the classifier in conceptualized phrase-mining-based concept mining and sequence tagging in our model. We utilize CRF++ v0.58 to train our model. 80\% of the dataset is used as the training set, 10\% as the development set, and the remaining 10\% as the test set.

\subsection{Publication of Datasets}

\begin{figure}
\centering
\includegraphics [width=0.45\textwidth]{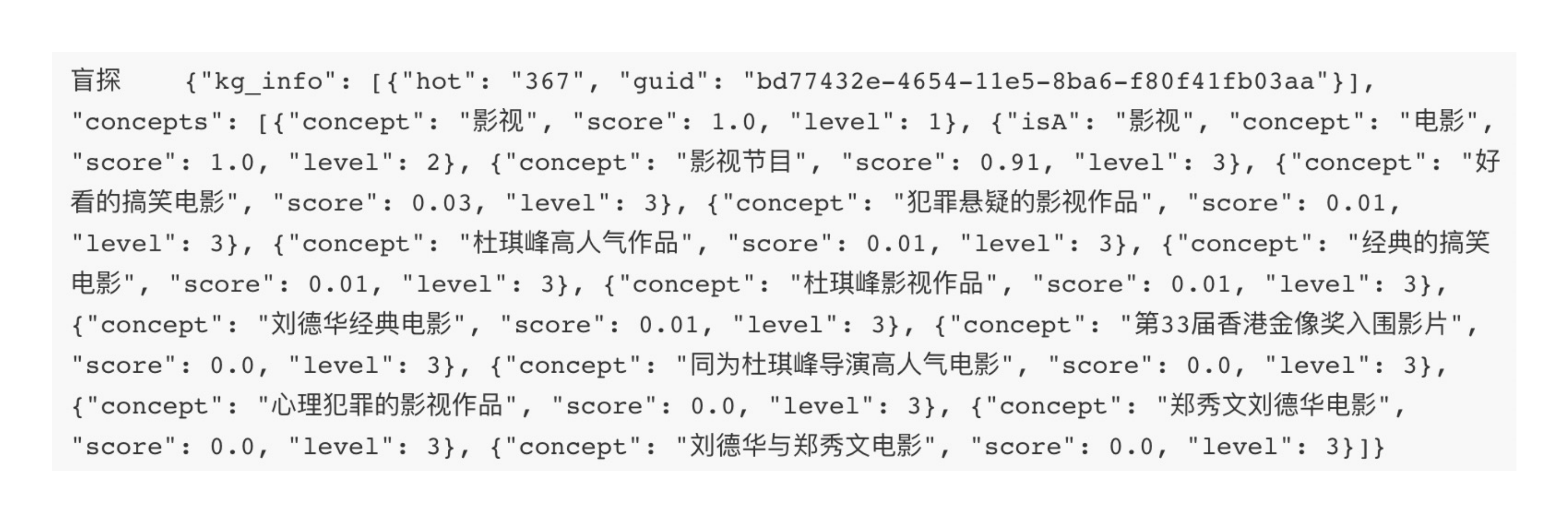}
\caption{Sample data in Alibaba Conceptual Graph.}
\label{sample}
\end{figure}

We have published our datasets for research purposes, and they can be accessed from Github\footnote{\url{https://github.com/alibaba-research/ConceptGraph}}.  Our dataset's major difference compared with Tencent's UCCG dataset is that 1) our dataset is bigger (50x); 2)  our dataset has concept possibilities; 3) our dataset contains long-tail concepts. Specifically, we have published the following open-source data: 
\begin{itemize}
\item  \textbf{The ACG dataset}. It is used to evaluate the performance of our approach for concept mining, and it contains 490,000 instances, as shown in Figure \ref{sample}. It is the largest conceptual graph dataset in Chinese. 

\item \textbf{The query tagging dataset}. It is used to evaluate the query tagging accuracy of AliCG, and it contains 10,000 queries with concept tags. 
 
\item  \textbf{The seed concept patterns for bootstrapping-based concept mining}. It contains the seed string patterns that we utilize for bootstrapping-based concept mining from queries.

\item  \textbf{Predefined level-1 and level-2 concept  list}. It contains more than 200 predefined concepts for taxonomy construction.

\end{itemize}

\subsection{Details of Entity Heat Estimation}

We introduce the details of entity heat estimation as we leveraged in confidence estimation in Sec. \ref{dynamic}.  The entity heat estimation algorithm calculates the popularity of an entity based on search behavior. We firstly parse user search logs then retrieval all the candidate entities from the knowledge graph via matching. We link those identified entities through various distances between entity mentions and entities. Finally, we sum the user search values and normalize heat values. We update the heat score of entities in the knowledge graph every day.

\subsection{Details of Entity Recommendation}
We utilized text rewriting and conceptualized pretraining for entity recommendations. For a simple query with entities, we utilize heterogeneous graph embedding to retrieve related entities.   For those complex queries with little entities, we propose a deep collaborative matching model to get related entities.   Then we rank those entities by various strategies, including type filtering, learning to rank,  and click-through rate estimation. 

 \begin{table}
 \caption{Part of the query-concept samples.}
\label{case}
    \small
    \centering
\begin{tabular}{p{3.7cm}p{3.7cm}}
 
\hline
Query&Concept\\
\hline
What are the Hangzhou special local product
 (\begin{CJK}{UTF8}{gbsn}杭州的特产有哪些\end{CJK}) & Hangzhou special local product (\begin{CJK}{UTF8}{gbsn}杭州特产\end{CJK})\\

Pancakes  cooking methods list (\begin{CJK}{UTF8}{gbsn}煎饼的做法大全\end{CJK}) & pancakes cooking methods (\begin{CJK}{UTF8}{gbsn}煎饼的做法\end{CJK}) \\
Which cars are cheap and fuel-efficient? (\begin{CJK}{UTF8}{gbsn}有什么便宜省油的车\end{CJK}) & cheap and fuel-efficient cars (\begin{CJK}{UTF8}{gbsn}便宜省油的车\end{CJK})  \\
Hangzhou famous snacks (\begin{CJK}{UTF8}{gbsn}杭州有名的小吃\end{CJK}) & Hangzhou snacks (\begin{CJK}{UTF8}{gbsn}杭州小吃\end{CJK}) \\
What are the symptoms of coronavirus pneumonia (\begin{CJK}{UTF8}{gbsn}冠状病毒肺炎有什么症状\end{CJK}) &symptoms of coronavirus pneumonia (\begin{CJK}{UTF8}{gbsn}冠状病毒肺炎症状\end{CJK}) \\
Latest fairy TV Series (\begin{CJK}{UTF8}{gbsn}最新仙侠电视剧\end{CJK}) & fairy TV Series (\begin{CJK}{UTF8}{gbsn}仙侠电视剧\end{CJK})\\
\hline
\end{tabular}
\end{table}

\subsection{Examples of Queries and Extracted Concepts}
Table \ref{case} lists a few examples of user queries, along with the concepts extracted by AliCG. The concepts are appropriate for summarizing the core user intention in the queries.

\end{document}